\documentclass[letterpaper]{article} 
\usepackage{aaai2026}  
\usepackage{times}  
\usepackage{helvet}  
\usepackage{courier}  
\usepackage[hyphens]{url}  
\usepackage{graphicx} 
\usepackage{natbib}  
\usepackage{caption} 
\usepackage{algorithm}
\usepackage{algorithmic}
\usepackage{acronym}
\usepackage{amsmath}
\usepackage{newfloat}
\usepackage{listings}
\DeclareCaptionStyle{ruled}{labelfont=normalfont,labelsep=colon,strut=off} 
\floatstyle{ruled}
\newfloat{listing}{tb}{lst}{}
\floatname{listing}{Listing}
\newacro{LLM}{Large Language Model}
\newacroplural{LLM}{Large Language Models}

\title{Mind the Gap: The Divergence Between Human and LLM-Generated Tasks}
\author{
    Yi-Long Lu\thanks{Corresponding authors},
    Jiajun Song,
    Chunhui Zhang,
    Wei Wang\footnotemark[1]
}
\affiliations{
    State Key Laboratory of General Artificial Intelligence, BIGAI, Beijing, China\\
    6yilong@gmail.com,
    zhangchunhui@bigai.ai,
    songjiajun@bigai.ai,
    wangwei@bigai.ai
}

\begin{document}

\maketitle

\begin{abstract}
    Humans constantly generate a diverse range of tasks guided by internal motivations. While generative agents powered by large language models (LLMs) aim to simulate this complex behavior, it remains uncertain whether they operate on similar cognitive principles. To address this, we conducted a task-generation experiment comparing human responses with those of an LLM (GPT-4o). We find that human task generation is consistently influenced by psychological drivers, including personal values (e.g., Openness to Change) and cognitive style. 
    Even when these psychological drivers are explicitly provided to the LLM, it fails to reflect the corresponding behavioral patterns. 
    They produce tasks that are markedly less social, less physical, and thematically biased toward abstraction. Interestingly, while the LLM's tasks were perceived as more fun and novel, 
    this highlights a disconnect between its linguistic proficiency and its capacity to generate human-like, embodied goals.
    We conclude that there is a core gap between the value-driven, embodied nature of human cognition and the statistical patterns of LLMs, highlighting the necessity of incorporating intrinsic motivation and physical grounding into the design of more human-aligned agents.
\end{abstract}

\begin{links}
    \link{Code}{https://github.com/Yilong-Lu/Mind_the_Gap}
\end{links}

\begin{figure*}[t]
    \centering
    \includegraphics[width=0.99\textwidth]{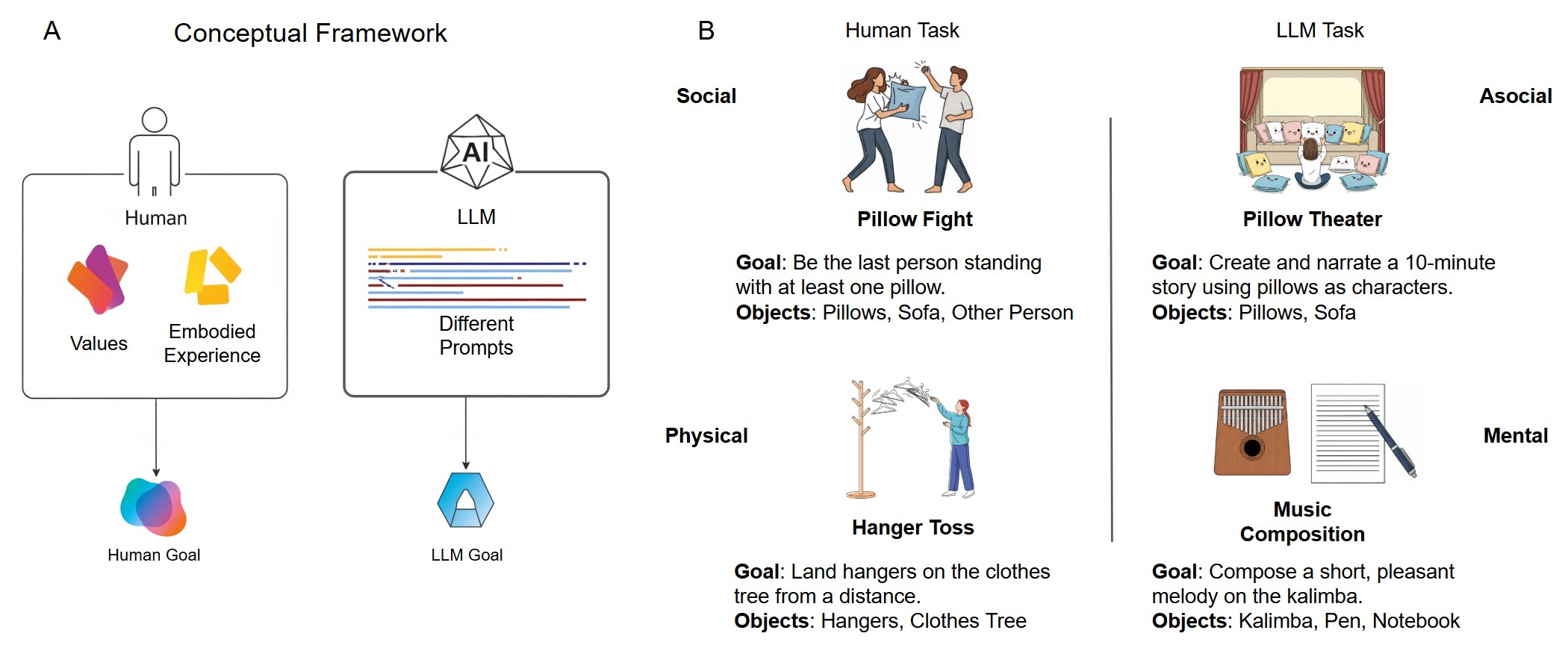}
    \caption{Overview of the study. (A) Conceptual framework. Human-generated tasks are typically driven by intrinsic motivation and grounded in embodied experience. In contrast, LLM-generated tasks are produced based on input prompts and their training data, which may result in a fundamental gap between the two. (B) Illustration of the thematic and embodiment gap of the tasks. Human-generated tasks tend to be more social and physically engaging, while LLM-generated tasks are less socially oriented and more abstract or cognitively focused.}
    \label{fig1}
\end{figure*}

\section{Introduction}
Humans can generate an infinite variety of goals to guide their behavior and enrich their daily lives, often infused with a distinct personal touch. This ability of autonomous task generation plays a central role in human cognition, shaping how individuals adapt to and interact with the world \cite{chuPraiseFollyFlexible2024,molinaroGoalcentricOutlookLearning2023}. 
Recent advances in \acp{LLM} have enabled generative agents that simulate human activity in virtual worlds with striking autonomy \cite{parkGenerativeAgentsInteractive2023,yangOASISOpenAgent2024}. 
This has spurred the development of benchmarks like TaskBench \cite{shen2024taskbench} and AgentGen \cite{hu2024agentgen}, which evaluate the planning and task automation capabilities of these models. However, these performance-focused evaluations often overlook a more fundamental question: does this simulated autonomy capture the deep cognitive mechanisms that drive human behavior, or is it merely a sophisticated mimicry of surface patterns? 
Can LLMs, trained on vast corpora of text, truly replicate the diversity, personal flair, and intrinsic motivation that characterize human goal-setting?

Answering this question requires shifting from performance benchmarks to an analysis of the internal drivers behind task generation. Human goal-setting is not an unconstrained, probabilistic generation of action tokens. It is shaped by two core pillars of cognition.

The first is a system of \textbf{value-driven} motivation. Personal values are stable, trans-situational goals that serve as guiding principles in life \cite{sagivPersonalValuesHuman2017, schwartzUniversalsContentStructure1992,sagivHowValuesAffect2021}. They provide the core motive force for human actions, shaping what we find desirable or important. The value dimension of \textit{openness to change} versus \textit{conservation}, for example, reflects a deep-seated conflict between the pursuit of novelty and stimulation versus a preference for tradition and security \cite{schwartzUniversalsContentStructure1992, broschNotMyFuture2018}, directly influencing creative and exploratory behaviors. These values shape not only what we do, but how we prioritize among competing possibilities. 

Second, human goals are shaped by \textbf{embodied experience}, that is, by the sensorimotor constraints of the body and accumulated interactions with the physical and social world \cite{varela2017embodied,foglia2013embodied,xiang2023language}. Humans naturally understand objects not just by semantic labels, but by how they can be physically manipulated and used in context. This grounding shapes how we imagine, evaluate, and select possible actions in real-world settings.

These two factors are largely absent from current LLMs. While recent LLM-based generative agents can exhibit behaviorally rich outputs, it remains unclear whether they possess mechanisms for prioritizing goals based on internal needs or for evaluating the physical feasibility and affordances of objects. Although emerging approaches model goals as procedural programs \cite{davidsonGoalsRewardProducingPrograms2024}, the crucial influence of these deep psychological and embodied factors remains largely overlooked.

\subsection{Research Objectives and Hypotheses}
To investigate these issues, we designed a novel text-based task generation paradigm that elicits unconstrained, intrinsically motivated responses from both humans and LLMs (Fig. \ref{fig1}). Our approach proceeds in two steps.

First, we examine how human task generation reflects underlying personal factors, establishing a behavioral signature of autonomous task generation\footnote{Our hypotheses were pre-registered at AsPredicted platform https://aspredicted.org/t3yb-wcd3.pdf}. We hypothesize that if human task generation is indeed \textbf{value-driven}, then participants' personal values (e.g., openness to change) should systematically predict key attributes of their generated tasks, such as their creativity and novelty.
Furthermore, we investigate how individuals strategically adapt to environmental complexity by examining their \textbf{cognitive style} \cite{sagivStructureFreedomCreativity2010}. We conceptualize cognitive style as the mechanism through which value-driven goals are pursued under varying cognitive loads. This aligns with dual-process theories distinguishing between deliberate, rule-based processing (systematic) and rapid, associative processing (intuitive) \cite{evans2013dual,sagivNotAllGreat2014}. We reasoned that individuals relying on systematic processing would be more sensitive to the increased cognitive demand of complex environments, affecting the diversity and nature of their generated tasks.

Second, we compare this human baseline to the output of a widely used LLM (GPT-4o), conditioned on the same value. This allows us to evaluate whether such models can replicate the psychological signatures of human goal-setting or whether fundamental differences remain, particularly in terms of motivational grounding and embodied realism.

\subsection{Contributions}
Our study yields the following key contributions:
\begin{enumerate}
    \item \textbf{Human Signatures of Autonomous Goal-Setting}: We provide behavioral evidence that human task generation is systematically shaped by personal values and cognitive style, supporting a value-driven account of autonomous task generation.
    \item \textbf{LLM Outputs Lack Value-Driven and Embodied Signatures}: Despite being prompted with individual human profiles (including personal values and cognitive styles), the LLM fails to exhibit core behavioral signatures of human goal generation.
    \begin{itemize}
        \item \textit{No Induction of Human-Like Behavior}: Conditioning on value-related inputs does not lead the LLM to generate behavior that mirrors human patterns of goal-setting. This suggests a lack of internal mechanisms for value prioritization or motivational grounding.
        \item \textit{Thematic and Embodiment Gap}: The LLM’s outputs show a strong thematic bias, favoring abstract over social or physical activities, and result in tasks perceived as more mentally demanding and less physically embodied than those generated by humans.
    \end{itemize}
\end{enumerate}

\section{Related Work}
Our research connects two domains: the psychology of autonomous goal-setting in humans and the challenge of simulating this behavior with LLMs, particularly concerning their psychological fidelity and embodied grounding.

\subsection{Autonomous Goal-Setting and Individual Differences}
Classic goal-setting theory has established how specific, difficult goals can enhance performance. However, much of human life is guided by autonomously generated goals in open-ended environments. Recent computational work has begun to model this process, for instance, by representing goals as programs in a domain-specific language \cite{davidsonGoalsRewardProducingPrograms2024}. While promising, these approaches often operate within constrained task spaces and do not account for the diverse psychological factors that drive human choice.

Human goals are not merely procedural; they are expressions of stable and diverse motivations. A large body of psychological research shows that personal values, trans-situational life principles like security, achievement, or benevolence, systematically guide attitudes and behavior \cite{sagivHowValuesAffect2021,sagivPersonalValuesHuman2017,kasofValuesCreativity2007}. Similarly, cognitive styles, such as the preference for systematic versus intuitive thinking, shape how individuals approach problems and adapt to environmental complexity \cite{evans2013dual,sagivNotAllGreat2014}. We incorporate both factors to define a human behavioral baseline for autonomous goal generation.

\subsection{LLMs as Cognitive Simulators: Capabilities and Limitations}
Recent advances have positioned LLMs as powerful tools for simulating human cognition and behavior. On one hand, their capabilities in task automation and planning are rapidly improving, pushing the boundaries of what models can execute in complex environments \cite{parkGenerativeAgentsInteractive2023, yang2025oasisopenagentsocial,shen2024taskbench, hu2024agentgen}. On the other hand, studies show LLMs can even replicate nuanced human social behaviors, including prosocial irrationality and cognitive biases, suggesting a surprising consistency with human decision-making patterns \cite{liu2025exploringprosocialirrationalityllm}. These capabilities have led to growing interest in using LLMs as cognitive models to explore how humans think, decide, and act.

Despite these capabilities, researchers have identified significant gaps in the psychological plausibility of these models. While LLMs can mimic surface-level reasoning and planning, they often lack deeper cognitive mechanisms such as emotion, causality, physical dynamics and social cognition \cite{hu2025reevaluatingtheorymindevaluation, binz2023using, ullman2024illusionillusionvisionlanguagemodels}. More fundamentally, because they are trained purely on text, they lack the sensorimotor grounding that shapes human goal formation. Embodied cognition theories emphasize that human goals and reasoning are deeply shaped by our physical experiences and social embeddedness \cite{varela2017embodied}. This suggests that generating goals rooted in physical interaction and embodied experience is a critical test case for the fidelity of LLM simulators—a test we conduct in this study.

\section{Methods}
We conducted two experiments to identify the unique signatures of human task generation by contrasting it with the output of a \ac{LLM}. Experiment 1 elicited tasks from both human participants and an \ac{LLM}. For the human baseline, we measured core psychological constructs hypothesized to drive autonomous behavior: stable motivational drivers \cite[personal values, via PVQ21, ][]{schwartzRepositorySchwartzValue2021}, cognitive strategies for navigating complexity \cite[thinking style, via TWS, ][]{sagivStructureFreedomCreativity2010}, along with emotional states as controls. Experiment 2 then involved an independent sample of human raters who evaluated key attributes of the generated tasks, enabling a direct comparison between human and AI outputs.

\begin{figure*}[t]
    \centering
    \includegraphics[width=0.9\textwidth]{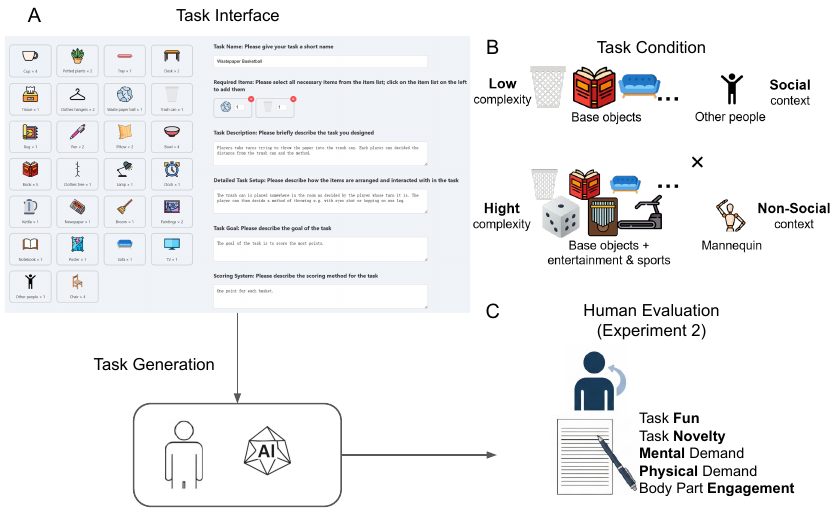} 
    \caption{Experimental interface and procedure. 
(A) Text-based task generation interface. Participants were asked to generate tasks using a given set of room items. They were instructed to report the task name, required items, detailed setup, goals, and scoring rules. 
(B) Experimental conditions. Participants were randomly assigned to one of four room scenarios, varying in environmental complexity (high vs. low) and social context (presence vs. absence of other people), sampled from a virtual simulation platform. 
(C) Task evaluation phase. Independent raters assessed both human- and LLM-generated tasks across multiple dimensions, including Fun, Novelty, Mental Demand, and Physical Demand.
}

    \label{fig_exp}
\end{figure*}

\subsection{Experiment 1: Human and AI Task Generation}
\subsubsection{Participants}
A total of 180 participants were recruited via the Prolific platform
(https://www.prolific.com/). 4 participants were excluded for meaningless task response, resulted in 176 participants in total (110 female, 66 male; age range: 18-65). All participants were fluent in English and provided informed consent
prior to the experiment. Participants were paid approximately £6.4/h for their participation.

\subsubsection{Stimuli and Design}
The experiment employed a 2 (environmental context: high vs. low complexity) $\times$ 2 (social context: social vs. non-social) between-subjects factorial design (Fig. \ref{fig_exp}B).
The \textbf{environmental context} was manipulated to vary the creative constraints. In the \textit{high-complexity} condition, participants were presented with a list of 32 common living room items. To increase the cognitive demand for task generation in the \textit{low-complexity} condition, we removed seven items with high entertainment affordances (e.g., Dice, Basketball), leaving 25 common items (e.g., Pens, Chairs).

The \textbf{social context} was manipulated to examine the influence of social motivation and embodied understanding. The \textit{social} condition included a ``person'' icon in the item list to enable interaction-based goals. The \textit{non-social} condition replaced ``person'' with ``mannequin'' to present a human-like form devoid of social agency. This manipulation isolates goal generation cued by social presence from that cued merely by a humanoid shape.

All object stimuli were selected from the room asset library of the virtual simulation based on \textit{Unreal Engine
 5} platform for ecological validity.

\subsubsection{Psychological Measures}
Personal value priorities were measured with the PVQ21 \cite{schwartzRepositorySchwartzValue2021}. Following established procedures, we computed scores for two orthogonal dimensions: openness to change (openness vs. conservation) and self-interest (self-enhancement vs. self-transcendence). Thinking style was measured with the Thinking and Working Style (TWS) Questionnaire \cite{sagivStructureFreedomCreativity2010}, which places individuals on a continuum from intuitive to systematic thinking. We administered the sociability facet of the Big Five Inventory–2 (BFI-2) \cite{sotoNextBigFive2017} as a targeted measure of the propensity for social interaction. The emotional valence and arousal for the day were also recorded to control participants' short-term mental states.

\subsubsection{Procedure}
After completing the psychological scales, participants were randomly assigned to one of the four conditions. They were shown a list of objects and instructed: "Please imagine that you are in a real room with several items in it. You need to use these items to pass the time." For each task, participants provided a title, a list of used objects, a description, task setup, goal, and scoring system (Fig. \ref{fig_exp}A). They reported the mental effort and perceived difficulty for each task generated. Participants were required to generate at least three tasks before choosing to conclude the experiment.

\subsubsection{LLM Implementation}
We implemented two GPT-4o conditions to test whether the model's deviation from human goal-setting stems from a simple information deficit or a fundamental mechanism differences. In the \textit{raw} condition, the model received only the task prompt, serving as a baseline for its default behavior. In the \textit{matched} condition, each model was given the full psychological profile of a corresponding human participant, including values and cognitive style. This condition directly tested whether the model could translate these psychological constructs into value-driven behavior when provided as semantic data, or whether its lack of internal motivation and embodied grounding imposes a deeper constraint. Both LLMs operated at a temperature of 1.0 to encourage diverse outputs.

\subsection{Experiment 2: Task Attribute Evaluation}
\subsubsection{Participants}
77 participants were recruited via an online platform. Two were excluded for random responding, leaving 75 evaluators (35 male, 30 female; age range: 18-31). They were compensated for their time.

\subsubsection{Stimuli and Procedure}
The stimulus pool comprised two task sets. To capture the full spectrum of human creativity, we included all 575 valid tasks generated by human participants. For a focused comparison, we compared this to the output from the matched GPT condition, including the first task generated for each of the 176 unique participant profiles. This ensures a representative, non-redundant sample of the model's performance when conditioned on individual data. Tasks from the raw GPT model were reserved for computational analyses (e.g., topic modeling) but excluded from human rating to streamline the evaluation process. Each of the 75 evaluators then rated a randomized subset of 50 tasks, yielding approximately five independent ratings per task. The evaluation was conducted in a blind manner.

\subsubsection{Evaluation Metrics}
Evaluators rated each task on a 0–10 scale across dimensions capturing creativity (Fun, Novelty) and embodiment (Overall Difficulty, Mental Demand, Physical Demand, Body Part Engagement). These metrics provided a quantitative basis for comparing human- and AI-generated goals (Fig. \ref{fig_exp}C).

We assessed inter-rater reliability using the Intraclass Correlation Coefficient \cite[ICC, ][]{koo2016guideline} based on a two-way random-effects model. Given the variable number of raters per item, we fitted a linear mixed-effects model to estimate variance components and calculated ICC(2,k) using the average number of raters (k). The ICCs for task dimensions rated by participants ranged from 0.58 to 0.90, indicating moderate to good reliability \cite{koo2016guideline}. Given that dimensions such as Fun involve considerable subjectivity, these values suggest that the measurements were acceptably reliable.

\subsection{Data Analysis}
\paragraph{Task Content Analysis.}
To identify emergent themes in the generated tasks, we performed topic modeling using BERTopic with embedding model ``all-mpnet-base-v2'' \cite{grootendorstBERTopicNeuralTopic2022, reimersSentenceBERTSentenceEmbeddings2019} on the full corpus of 1718 tasks (575 human, 575 matched GPT, and 568 raw GPT tasks). To measure the creativity of each participant's output, we computed their task diversity score, defined as the mean cosine distance between the sentence-transformer embeddings of the tasks they generated.


\paragraph{Human Baseline Analysis.} To test our hypotheses about the link between psychological traits and task attributes, we specified a series of linear mixed-effects models \cite[LMMs, ][]{brauer2018linear} to predict key dependent variables: the rated attributes of Fun, Novelty, and their task content diversity.
We first specified a baseline model including only demographic variables and random participants intercepts. Then we specified a main-effects model by adding experimental conditions and psychological predictors like personal values and thinking style. Finally, we included the pre-specified TWS × Environmental Context interaction into the model. The models were compared using a likelihood ratio test \cite[LRT, ][]{moreira2003conditional} to determine if including the psychological predictors significantly improved model fit.

\paragraph{Human-LLM Comparison.} We used chi-square tests to compare the distribution of task topics and the frequency of multiplayer task generation. We used Mann-Whitney U tests \cite{mcknight2010mann} with false discovery rate (FDR) correction to compare the rated attributes of human- and AI-generated tasks.

\begin{figure*}[thbp]
    \centering
    \includegraphics[width=0.92\textwidth]{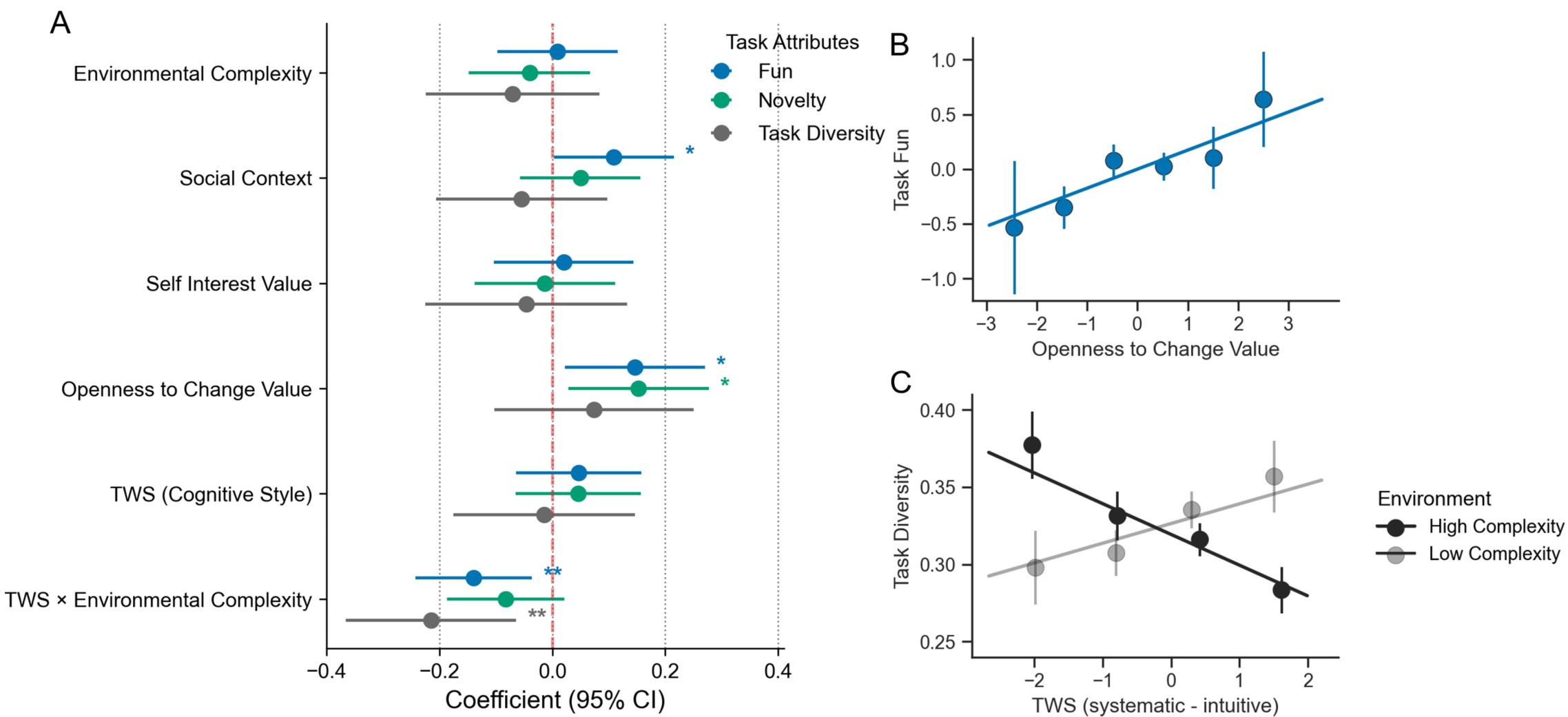}
    \caption{Personal values and cognitive style shape human goal generation.
    (A) Regression coefficients of key predictors on task attributes (Novelty, Fun, and Task Diversity). * indicates $p < 0.05$, ** indicates $p < 0.01$. 
    (B) Openness to Change values predict task fun: as Openness to Change increases, participants' tasks are rated as more enjoyable. 
    (C) Cognitive style (TWS) interacts with environmental complexity to predict task diversity: in high-complexity environments (black), individuals with intuitive styles (lower TWS) produce more diverse tasks. 
    Bars indicate 95\% confidence intervals.}
    \label{fig2}
\end{figure*}

\section{Results}

Our analyses proceed in two stages. First, we establish a behavioral signature of human goal generation, demonstrating its systematic link to personal values and environmental context. Second, we contrast this human baseline against the output of LLMs to reveal their gaps.

\subsection{The Signature of Human Goal Generation: Value-Driven and Environmentally Sensitive}
To establish a human baseline, we first tested how psychological traits predicted people's goals. Linear mixed-effects models confirmed that task generation was systematically linked to stable personal values (Fig. \ref{fig2}A). The value of Openness to Change significantly predicted higher ratings of both task novelty ($b = 0.152$, $t = 2.406$, $p = 0.017$) and fun (Fig. \ref{fig2}B, $b = 0.146$, $t = 2.316$, $p = 0.022$). In contrast, the self-interest value dimension showed no such relationship. This finding supports the core hypothesis that stable motivational principles guide autonomous goal-setting.

Behavior was not only driven by stable values but was also adapted to the environment. We found that the creativity of generated tasks depended on a significant interaction between an individual’s cognitive style (TWS) and the complexity of the environment (Fig. \ref{fig2}A). This pattern appeared in both subjective ratings of task fun (interaction of TWS and Environmental Context: $b = -0.140$, $t = -2.640$, $p = 0.009$), and in objective measures of task diversity (Fig. \ref{fig2}C, $b = -0.215$, $t = -2.809$, $p = 0.006$). Specifically, individuals with a more systematic thinking style generated more fun tasks in the simple environment, whereas those with an intuitive style thrived in the complex one. This suggests people flexibly deploy different cognitive strategies to match environmental demands.

\begin{figure*}[t]
    \centering
    \includegraphics[width=0.9\textwidth]{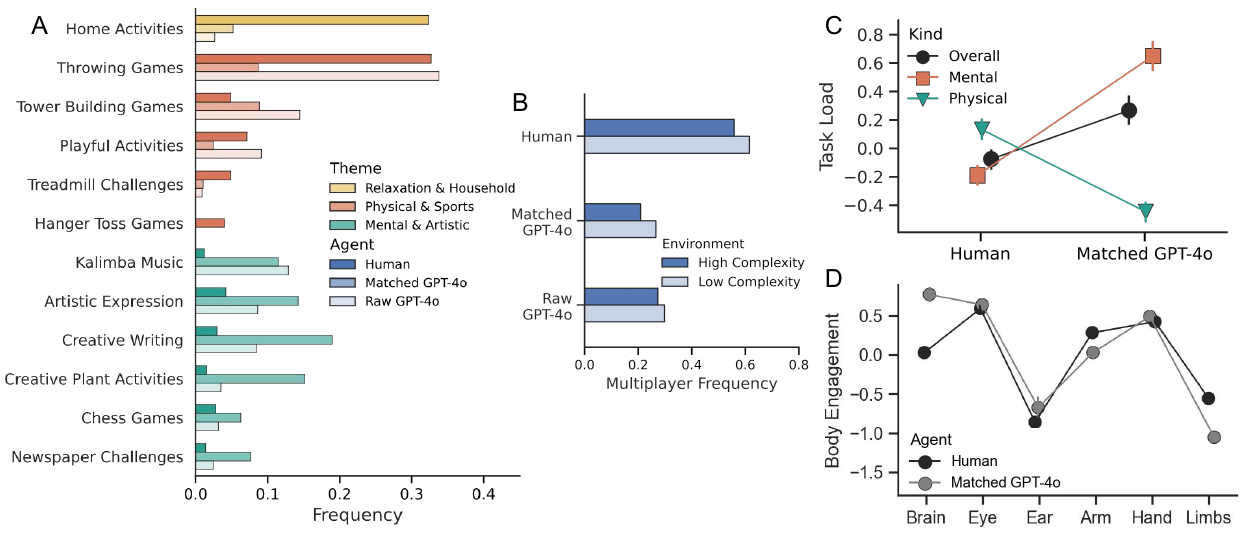} 
    \caption{Human and GPT generated systematically different goals. (A) Thematic distribution of human and LLM tasks. 1718 tasks generated by human and GPT were clustered into 12 topics and 3 themes: ``Physical \& Sports Activities", ``Relaxation \& Household Activities", and ``Mental \& Artistic Activities". (B) LLMs exhibit a lower propensity for social task generation. (C) LLM tasks are perceived as more mentally and less physically demanding. Black dots stands for the overall difficulty ratings of tasks. Orange and green dots stands for mental and physical load respectively. (D) Engagement ratings of different body parts.}
    \label{fig3}
    \end{figure*}

\subsection{The Human-LLM Gap: A Disembodied and Asocial Simulation}
Having established a human behavioral signature, we next compared it to the output of LLMs to identify structural differences. The comparison revealed systematic differences in task content and structure.

\subsubsection{Thematic Differences in Task Generation}
We directly compared the distribution of task topics between human participants and LLMs. Topic modeling identified three dominant themes across the dataset: Physical \& Sports, Relaxation \& Household, and Mental \& Artistic. The distribution of tasks across these themes differed significantly between humans and the LLMs ($\chi^2(4)=581.45$, $p<0.001$). Human participants generated a balanced mix of activities. In contrast, LLMs showed a bias toward abstract tasks like music and writing (Fig. \ref{fig3}A). 

This bias was most evident in the Matched GPT condition, where ``Mental \& Artistic'' tasks accounted for 74\% of its output. For example, music-related tasks (``kalimba music'') were rare in human responses (1\%) but common for both Raw GPT (13\%) and Matched GPT (11\%). This focus on the abstract resulted in the near-total neglect of the ``Relaxation \& Household Activities'' theme by the models (3\% for Raw GPT, 5\% for Matched GPT), which was a major component of human-generated goals (32\%). Consequently, common embodied tasks involving using everyday objects like hangers and treadmills were also almost absent from the LLM's responses.

\subsubsection{Limited Social Interaction in LLM Tasks}
Beyond thematic content, \acp{LLM} generated significantly fewer social tasks than humans in settings where another person was present ($\chi^2(2)=93.19$, $p<0.001$). In these scenarios, 58\% of human-generated tasks were multiplayer, with 83\% of participants creating at least one social goal. Humans proposed a range of interactive tasks, from simple conversation games like ``Guess Who I Am'' to competitive games like ``Pillow Fight''. In contrast, multiplayer tasks from the Raw and Matched GPT were far less common, accounting for only 29\% and 24\% of their outputs respectively (Fig. \ref{fig3}B).

\subsubsection{A Disembodiment Gap in Perceived Demands}
The thematic differences between human and LLM tasks were reflected in their perceived mental and physical demands (Fig. \ref{fig3}C). Tasks generated by the Matched GPT were rated as significantly more mentally demanding ($z = -11.64$, FDR-corrected $p < 0.001$) and less physically demanding ($z = -8.40$, $p_{\text{FDR}} < 0.001$) than those generated by humans. This aligns with the LLM's focus on abstract, ``Mental \& Artistic'' activities.

This mental-physical divide was also evident in the specific body parts required for task execution (Fig. \ref{fig3}D). GPT tasks demanded greater brain engagement ($z = -11.40$, $p_{\text{FDR}} < 0.001$) but significantly less involvement of the arms ($z = -5.43$, $p_{\text{FDR}} < 0.001$) and lower body ($z = -7.93$,  $p_{\text{FDR}} < 0.001$). Engagement of the eyes, ears, and hands did not differ between the two groups.

\subsubsection{The Paradox of Ungrounded Creativity}
\begin{figure}[t]
    \centering
    \includegraphics[width=0.45\textwidth]{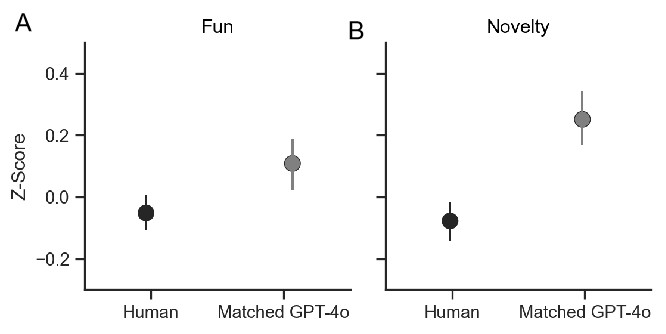} 
    \caption{LLMs generate tasks rated as more fun and novel.}
    \label{fig4}
\end{figure}

Despite their disconnection from physical and social experience, Matched GPT tasks were rated as more novel ($z = -5.63$,  $p_{\text{FDR}} < 0.001$) and more fun ($z = -2.73$,  $p_{\text{FDR}} = 0.010$) than those created by humans (Fig. \ref{fig4}). This highlights a key tradeoff: LLMs can generate text that seems creative and engaging, but they do so without grounding in embodied or socially situated experience.

\section{Discussion}
This study examined the cognitive basis of autonomous goal generation by comparing human behavior with \acp{LLM}. Our findings show that human-generated goals are systematically shaped by personal values and adapted to environmental complexity. In contrast, LLMs, even when prompted with personal profiles, exhibit a fundamental and predictable gap between their text-based simulation and the grounded nature of human cognition.

\subsection{Drivers of Human Goal Generation}
Individual differences shaped goal content. People high in openness to change generated more novel and fun tasks, consistent with prior work linking openness to exploratory behavior and intrinsic motivation \cite{deci2013intrinsic,mccrae1997personality,sagivHowValuesAffect2021}. Cognitive style also mattered: systematic thinkers generated less diverse tasks in complex environments, possibly due to higher cognitive load. These results align with theories of creativity and decision-making, which suggest that systematic thinkers prefer structured problem-solving, while intuitive thinkers are more adaptable in diverse settings \cite{evans2013dual,sagivNotAllGreat2014}.

\subsection{Explaining the Human-LLM Gap}
The systematic differences we observed point to two core mismatches.

First is the embodiment gap. Humans generated tasks involving physical interaction with common objects (e.g., hangers, pillows), reflecting sensorimotor knowledge of object use.  LLMs rarely produced such tasks, and their outputs were rated as less physically demanding and less involving of limbs. This supports the view that LLMs rely on text-based associations and lack access to intuitive physics or affordance perception \cite{lake2017building}.

Second is the value and motivation gap. Humans often proposed social and prosocial tasks, consistent with basic motivational drives like affiliation. In contrast, LLMs showed a lower propensity for social engagement and a strong bias toward abstract, mental tasks (e.g., music, writing). Crucially, providing the model with a user's psychological profile did not fix this. This finding suggests that describing values in a prompt is no substitute for an integrated value system that actively regulates goal selection.

\subsection{The Paradox of Aligned Creativity}
LLM-generated tasks were rated as more novel and fun. This highlights the unique nature of \ac{LLM} creativity. Unconstrained by physical feasibility, \acp{LLM} excel at combinatorial creativity, drawing on their vast training data to produce imaginative textual descriptions.

This natural tendency is likely amplified by current alignment methods like Reinforcement Learning from Human Feedback (RLHF) \cite{ouyang2022training, bai2022traininghelpfulharmlessassistant}. Such methods optimize for outputs that human raters perceive as immediately helpful or engaging. This reveals a fundamental tension between two objectives: developing an expert LLM skilled in creative or intellectual domains versus constructing one that mirrors the often mundane patterns of real human behavior. This optimization paradigm inherently discourages the generation of routine actions like resting or cleaning. The result is a model fine-tuned to produce ``ungrounded'' but highly-rated novelties, explaining both its high creativity scores and its thematic detachment from the reality of human experience.

\subsection{Future Directions and Limitations}

Our findings have two main implications. For AI research, they suggest that creating truly human-like autonomous agents requires more than scaling up \acp{LLM}. Addressing the embodiment and motivation gaps may require integrating world models, intrinsic reward systems, or sensorimotor learning \cite{matsuo2022deep}. For cognitive science, our task-based paradigm offers a scalable, flexible tool to study goal-setting in naturalistic contexts, beyond traditional lab constraints. It enables new comparisons between human cognition and machine outputs.

Several limitations of our study point toward promising future directions. 
First, \textbf{the text-based task generation paradigm} may not fully capture the richness of human goal-setting, which often involves multimodal, interactive, and context-sensitive elements. Future work could incorporate interactive 3D environments with more scenarios to enhance ecological validity. 
Second, due to the high cost of human evaluation, our study focused on a deep, theory-driven analysis of a \textbf{single architecture} (GPT-4o), which represented the state-of-the-art at the time of our pre-registration. This approach prioritizes depth over breadth, establishing a detailed cognitive baseline. Future work should expand these comparisons across a broader range of models to test the generality of our findings. Finally, the complexity of human-generated goals calls for new computational frameworks capable of modeling open-ended, value-guided behavior.

\section*{Acknowledgements}
This work was partially supported by the National Science Foundation of China under grant No. 32441106.

\bigskip
\bibliography{AutoTask.bib}

\end{document}